\documentclass[11pt]{article}

\usepackage[
  twocolumnmode,              
  shownumpages,               
  bgcolor={245,245,250},      
  braincolor={190,32,240},    
  citingstyle=authoryear,     
  bibliostyle=plainnat,       
  bibfile=refs                
]{brainlab}

\usepackage{microtype}
\usepackage{graphicx}
\usepackage{subfigure}
\usepackage{booktabs} 
\usepackage{hyperref}
\usepackage{wrapfig}
\usepackage{multirow}
\usepackage{cuted}
\usepackage{multicol}

\setbrainmeta{
  title={\texttt{WeightLoRA}: Keep Only Necessary Adapters},
  authors={
    Andrey Veprikov\textsuperscript{1, 2}, Vladimir Solodkin\textsuperscript{1}, Alexander Zyl\textsuperscript{2}, Andrey Savchenko\textsuperscript{2}, Aleksandr Beznosikov\textsuperscript{1, 3}
  },
  affiliations={
    \textsuperscript{1}Basic Research of Artificial Intelligence Laboratory (BRAIn Lab) \\
    \textsuperscript{2}SB AI Lab \\
    \textsuperscript{3}Innopolis University \\
  },
  abstract={
    The widespread utilization of language models in modern applications is inconceivable without Parameter-Efficient Fine-Tuning techniques, such as low-rank adaptation ($\texttt{LoRA}$), which adds trainable adapters to selected layers. Although $\texttt{LoRA}$ may obtain accurate solutions, it requires significant memory to train large models and intuition on which layers to add adapters. In this paper, we propose a novel method, $\texttt{WeightLoRA}$, which overcomes this issue by adaptive selection of the most critical $\texttt{LoRA}$ heads throughout the optimization process. As a result, we can significantly reduce the number of trainable parameters while maintaining the capability to obtain consistent or even superior metric values. We conduct experiments for a series of competitive benchmarks and DeBERTa, BART, and Llama models, comparing our method with different adaptive approaches. The experimental results demonstrate the efficacy of $\texttt{WeightLoRA}$ and the superior performance of $\texttt{WeightLoRA+}$ in almost all cases.
  },
}

\begin{document}
\begin{mainpart}
\section{Introduction}
\label{sec:intro}
The emergence of Large Language Models (LLMs) has transformed numerous domains, from natural language processing to code generation~\cite{dubey2024llama,he2020deberta}. 
Despite the undebatable performance, their fine-tuning remains computationally intensive and memory-demanding. Parameter-Efficient Fine-Tuning (PEFT)~\cite{ding2023parameter,han2024parameter}, such as Low-Rank Adaptation (\texttt{LoRA})~\cite{hu2021lora}, has significantly alleviated these challenges by reducing the number of trainable parameters while maintaining competitive accuracy. 

However, \texttt{LoRA} still faces inherent limitations, including substantial memory requirements for its rank-specific adapters and the challenge of selecting appropriate hyperparameters, particularly the rank and position of the adapters, 
which significantly influence performance and resource consumption~\cite{liu2024alora,zhou2024lora}. Over the years, several modifications of \texttt{LoRA} have been introduced to deal with its disadvantages and improve efficiency~\cite{liu2024alora,liu2024dora,zhang2023adalora,zhou2024lora}. Although these modifications are generally effective in addressing certain deficiencies, empirical evidence shows that, for general staging applications, the conventional \texttt{LoRA} approach is often more advantageous due to the complexity inherent in the configuration of specific methods (see Section \ref{sec:exp}). Details about various \texttt{LoRA} modifications are presented in Section \ref{sec:related}.

In this paper, we 
propose \texttt{WeightLoRA} (Figure~\ref{fig:LoRA_vs_WeightLoRA}), a novel approach to fine-tuning LLMs that adaptively optimizes and selects \texttt{LoRA} adapters based on their contribution to model performance. 
Using an alternating minimization algorithm~\cite{karlin2018alternating}, 
we managed to keep only the top-$K$ adapters with the highest impact on fine-tuning. Our dual-phase approach significantly reduces the number of trainable parameters while maintaining high performance, making \texttt{WeightLoRA} a scalable solution for fine-tuning LLMs in resource-constrained environments. To evaluate the effectiveness of \texttt{WeightLoRA}, we conducted extensive experiments using natural language processing~\cite{wang2018glue}, question answering~\cite{rajpurkar2016squad}, and natural language
generation~\cite{narayan2018don} benchmarks with state-of-the-art low-budget LLMs, including DeBERTaV3-base~\cite{he2021debertav3}, BART-large~\cite{lewis2019bart} and Llama3-7B \cite{grattafiori2024llama3herdmodels}.

\begin{figure*}[t]
        \centering        
        \includegraphics[width=1.0\linewidth]{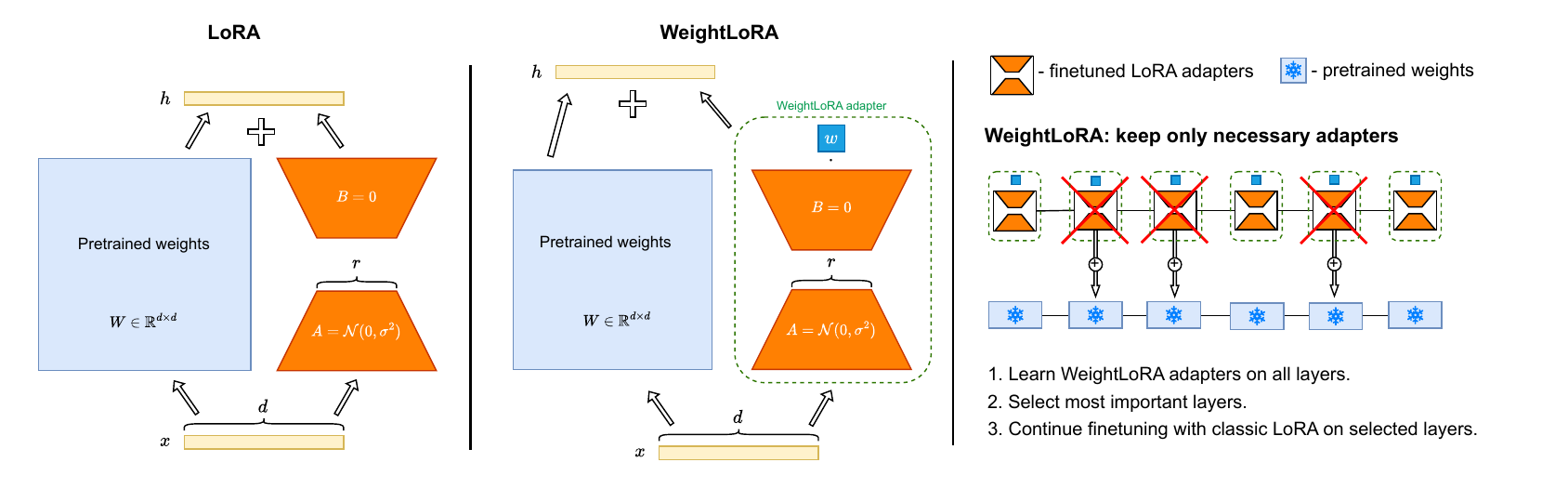}
        \caption{Comparison between \texttt{LoRA} (left) and the proposed \texttt{WeightLoRA} framework (right). The core idea of \texttt{WeightLoRA} is to add weights to the \texttt{LoRA} adapters, choose the most important ones, and train only this small subset.}        \label{fig:LoRA_vs_WeightLoRA}    
\end{figure*}

\paragraph{Our Contribution}\hfill

$\bullet$ We introduce \texttt{WeightLoRA}, a novel PEFT method that assigns a trainable weight to each adapter and selects the most important \texttt{LoRA} adapters with the highest weights during the training process. As a result, the number of trainable parameters is significantly reduced while maintaining high performance.

$\bullet$ We introduce \texttt{WeightLoRA+}, which advances the concept of \texttt{WeightLoRA} through an adaptive increase of the rank for the selected adapters. 

$\bullet$ We compare our approach with the original \texttt{LoRA} method and its various modifications on the classical fine-tuning benchmarks.

\section{Related Work}
\label{sec:related}

\textbf{Low-Rank Adaptation (LoRA).}
Traditional LoRA~\cite{hu2021lora} is one of the first successful applications of PEFT to LLMs. It avoids whole supervised fine-tuning by utilizing low-rank matrix decomposition to update a subset of model parameters. Unfortunately, \texttt{LoRA} requires careful selection of hyperparameters, namely, the ranks of adapters at each layer. As a result, constant rank is used in practice, which leads to a relatively high number of trainable parameters. Moreover, the accuracy of the obtained data may be lower than that of the whole fine-tuning. Numerous modifications have been proposed to overcome these shortcomings. Let us discuss some of them here. 

Several techniques utilize weight sharing to reduce the number of trainable weights without a decrease in training time. For example, \texttt{VeRA}~\cite{kopiczko2023vera} employs one pair of low-rank matrices shared across all layers and learns only small scaling vectors. \texttt{Tied-LoRA}~\cite{renduchintala2023tied} used selective training and weight-tying of low-rank matrices. However, such techniques may significantly decrease accuracy, especially if the ranks are very small.

Another way to share the parameters across different modules and layers is \texttt{VB-LoRA}~\cite{li2024vb}. Inspired by a ``divide-and-share'' paradigm, it combines rank-one decomposition with a shared vector bank, dividing vectors into sub-vectors for parameter sharing across layers and modules. This design and a differentiable top-k selection mechanism enable task-specific tuning with minimal storage overhead. \texttt{VB-LoRA} demonstrates compatible performance across different tasks but requires careful tuning of its hyperparameters. 

\texttt{DoRA} (Weight-Decomposed \texttt{LoRA})~\cite{liu2024dora} decomposes the LLM's weight into magnitude and direction parts. While it achieves competitive performance, it introduces an overhead in training time, which can be undesirable. Another modification, \texttt{NoRA}~\cite{lin2024nora} uses singular value decomposition (SVD) and a special dual structure, where the inner layer introduces fine-grained adjustments tailored to different tasks, and the outer layer provides a low-rank foundation by keeping its weights fixed. \texttt{MiLoRA}~\cite{wang2024milora} performs SVD and trains only the minor singular components. 

\textbf{Dynamic Rank Allocation.} Another recent trend in the literature is to design methods that dynamically choose \texttt{LoRA}'s parameters. This approach can be applied in a variety of ways. For instance, \texttt{AdaLoRA}~\cite{zhang2023adalora} performs adaptive rank selection during training using SVD to update weight matrices. As a result, it tries to better distribute the trainable weights over the Transformer layers. 

FlexLoRA \cite{wei2024flexora}, on the other hand, also utilizes the SVD approach, but in a completely federated setting. The algorithm sends truncated decompositions to each client based on their computational budget, allowing for dynamic resource-based rank allocation.

At the same time, IncreLoRA \cite{zhang2023increlora} increases ranks dynamically based on the importance scores computed as a smoothed quantified average of the scores of all parameters of the update matrix. This method also adds regularization terms that force these matrices to be orthogonal and introduces several additional parameters for storage.

Another model, Dynamic \texttt{LoRA}~\cite{mao2024dora}, removes adapters based on their importance to specific tasks during training. It also treats high-rank layers as a combination of single-rank components.

\texttt{LoRA}-drop~\cite{zhou2024lora} evaluates the importance of \texttt{LoRA} based on the product of the \texttt{LoRA} parameter and hidden state, dropping the adapters with small output. It achieves nearly the same accuracy as the original \texttt{LoRA} while reducing the number of trainable parameters. 

Allocating \texttt{LoRA} (\texttt{ALoRA})~\cite{liu2024alora} utilizes the diagonal matrix of gate units for each rank not greater than the initial \texttt{LoRA} rank. During training, the rank gates are set to zero for ranks with negative contributions, and the pruned rank budget is added to other weights. As a result, it can increase the accuracy of fine-tuning for a fixed memory budget.

\texttt{MELoRA}~\cite{ren2024melora} freezes the original pretrained weights and trains a group of mini \texttt{LoRA}s with very few parameters. Based on significant diversity among mini \texttt{LoRA}s, \texttt{MELoRA} promotes better generalization ability with reduced memory usage.

Despite the variety of existing approaches, the extant literature does not address the dynamic adapter selection option in its most direct variant, namely the use of trainable weights.

\section{\texttt{WeightLoRA} Framework}
In this section, we introduce \texttt{WeightLoRA}, a novel approach to parameter-efficient fine-tuning. To provide the foundation of our method, we begin with an illustrative example. 
\subsection{Motivating Example}
In this example, we explore how much adapters can differ from each other in terms of the need for tuning. It turns out that only a small subset of adapters require intensive training. 
More formally, let $f$ be a model with pretrained weights $\mathcal{W} = \{W^i\}_{i = 1}^n$ and adapters $\Delta \mathcal{W} = \{\Delta W^i\}_{i=1}^n$, where $i$ denotes the layer index and $n$ is the total number of layers. We are interested in solving the following problem: 
\begin{align}
\label{eq:motivex_1}
    \min_{\Delta \mathcal{W}} \left\{ f(W^1 + \Delta W^1, \ldots, W^n + \Delta W^n) \right\},
\end{align}
which is in our case equivalent to
\begin{align*}
    \min_{ \mathcal{A}, \mathcal{B}} \left\{ f(W^1 + \alpha_1 A^1B^1, \ldots, W^n + \alpha_n A^nB^n) \right\},
\end{align*}
where $\mathcal{A}, \mathcal{B}$ are the sets of all \texttt{LoRA} adapters (see details in Appendix \ref{sec:lora_notation}) and $\alpha = (\alpha_1, \ldots, \alpha_n)^T \in \mathbb{R}^n$ is an arbitrary fixed vector. Using the Taylor expansion of the function $f$ at the point $\alpha$, we obtain that:
\begin{align*}
     f(W^1 &+ \alpha_1A^1B^1, \ldots, W^n + \alpha_1A^nB^n) \approx f(\mathcal{W})\\ &+ \alpha_1\langle\nabla_{W^1} f(W^1, \ldots, W^n), A^1B^1\rangle + \ldots \\ &+ \alpha_n\langle\nabla_{W^n} f(W^1, \ldots, W^n), A^nB^n\rangle,
\end{align*}
for sufficiently small values of $\alpha$. That is, substituting this expression into the problem \eqref{eq:motivex_1} and omitting the unnecessary terms, we can rewrite it separately as follows:
\begin{equation}
    \label{eq:dotprods}
\begin{split}
    \eqref{eq:motivex_1} \Leftrightarrow& ~\alpha_1\min_{A^1B^1} \left\{\langle\nabla_{W^1} f(\mathcal{W}), A^1B^1\rangle \right\} + \ldots 
    \\&+ \alpha_n\min_{A^nB^n} \left\{\langle\nabla_{W^n} f(\mathcal{W}), A^nB^n\rangle \right\},
\end{split}
\end{equation}
In the general case, given this setting, all minima are distinct and have different magnitudes, illustrating potential differences in the tuning process. Consequently, different times may be required to optimize different $A^iB^i$ during training. Furthermore, it is possible that $A^iB^i$ does not need to be tuned at all, e.g., when the condition $\nabla_{W^i} f(W^1, \ldots, W^n)=0$ is met, which means that we do not need any fine-tuning for the layer $i$. 

To validate the theoretical considerations, we now show how the minima in \eqref{eq:dotprods} can deviate from each other in practice. Consider the DeBERTa \cite{he2021debertav3} model, which has $n=36$ self-attention layers.

We connect a single adapter $\Delta W^1$ to the first layer and fine-tune the model for several epochs. After that, we compute the full gradient $\nabla_{W^1} f(W^1, \ldots, W^n)$ and the dot product between the adapter and the gradient. Then we repeat this process for each adapter $\Delta W^i$ and layer $i$ on the same subset of the GLUE benchmark \cite{wang2018glue}. The results averaged over five random seeds are presented in Figure \ref{fig:adaptersdotprod}.

\begin{figure}[!h]
    \centering
    \includegraphics[width=1.\linewidth]{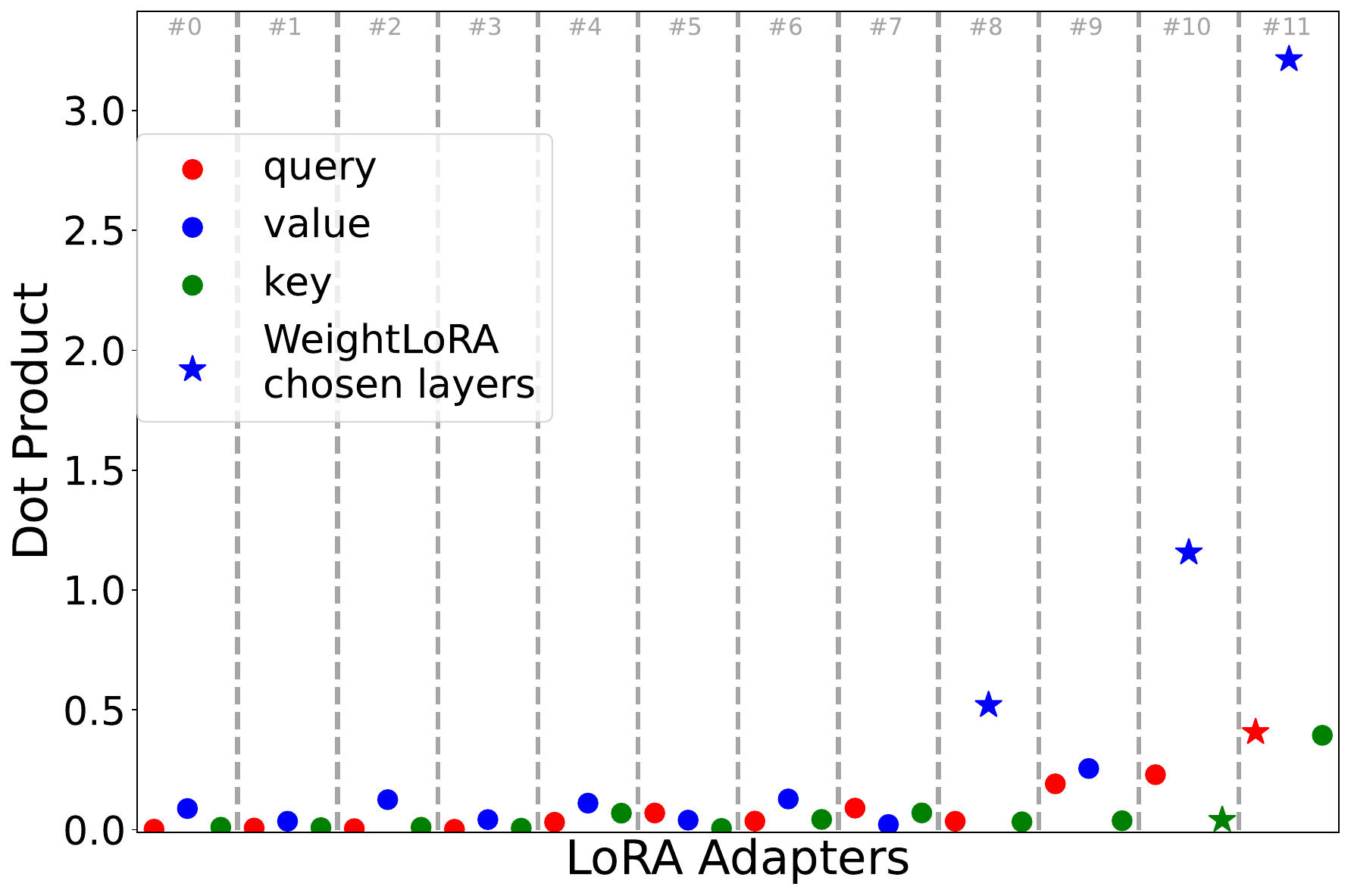}
    \caption{Comparison of the absolute values of the scalar products 
    $\langle\nabla_{W^i} f(\mathcal{W}), A^iB^i \rangle$ from \eqref{eq:dotprods} for all layers $i \in \{1, 2, ..., 36\}$. 
    The adapters selected through our \texttt{WeightLoRA} framework are starred.}
    \label{fig:adaptersdotprod}
\end{figure}

It is observable that the value adapters in layers $8, 10$, $11$, and the key and query adapters in layer $11$ have significantly higher absolute values of the scalar products. This implies that the learning process for these adapters to converge is considerably more time-consuming, and they are, to a certain extent, the sole determinant of the model's final performance.

For the sake of clarity, we emphasize that this is merely a motivating example, and our proposed method does not require the full gradient computation during training. Notably, this experiment can be fully extrapolated to models of any size and problem of any origin, but since it is very time-consuming, we leave it to the ablation study. 

\subsection{GPU Memory Experiment}

If we look at the relationship between GPU memory utilization and the number of enabled adapters (see Figure \ref{fig:memory}), we can expect a significant reduction in the memory required for model training by disabling several \texttt{LoRA} heads if competitive performance is maintained. 

For instance, let us assume that we have only one NVIDIA V100 16GB GPU available for training a DeBERTaV3-base model, which is a somewhat realistic setting in the context of fine-tuning a small (or large enough but quantized) model for a downstream task. According to the dashed line in Figure~\ref{fig:LoRA_vs_WeightLoRA}, this configuration allows for a maximum of twenty adapters, even considering a minimal $r = 1$ rank. If we spread these adapters across random layers, the probability of selecting precisely the \texttt{LoRA} heads that necessitate the most extensive fine-tuning is low. Consequently, the performance of the resulting model is substantially lower than that of classical \texttt{LoRA} (see Section \ref{appendix:ablation} for a detailed comparison). 

Conversely, if only five adapters are placed in optimal locations (also enabling the usage of higher ranks), gains in memory and time can be achieved without a loss in performance (see Section \ref{sec:exp}). That is, by keeping only the value adapters in layers $8, 10$, $11$ and the key and query adapters in layer $11$ enabled, the number of trainable parameters is reduced by $\sim 86\%$ if the rank $r = 8$ (see Table \ref{table:glue_all_adapters}).

If the rank and number of layers are sufficiently large, the benefit of fine-tuning only the adapters that require it becomes even more significant. This leads to the idea of choosing only "good" adapters and training them while disconnecting the rest. 

\begin{figure}[h]
    \centering
    \includegraphics[width=\linewidth]{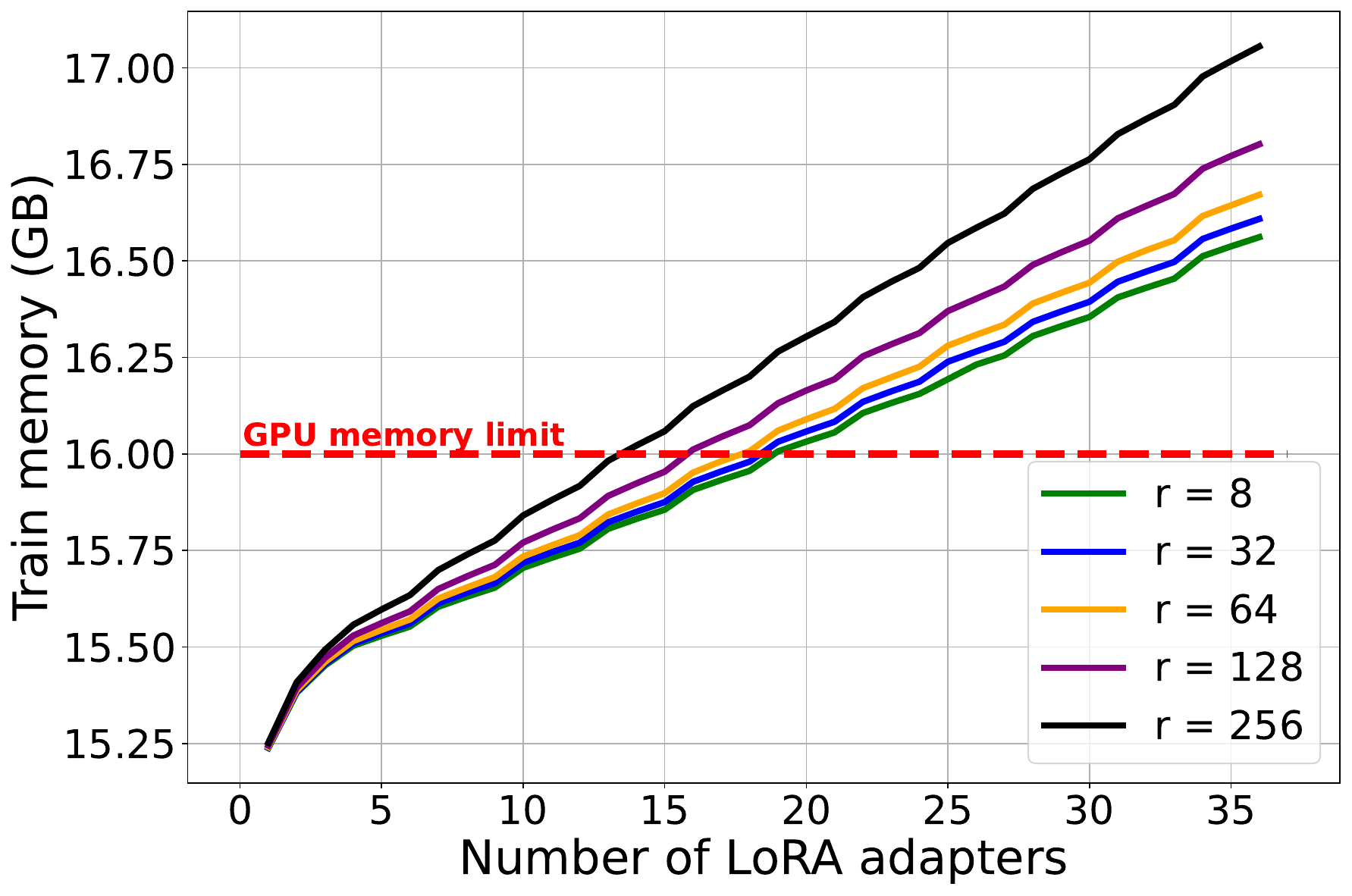}
    \caption{Dependence of the amount of required GPU memory on the number of connected \texttt{LoRA} adapters. The red dashed line indicates the memory capacity of the NVIDIA V100 16GB.}
    \label{fig:memory}
\end{figure}

\subsection{\texttt{WeightLoRA}}
Motivated by these considerations, we introduce the selection process for the most valuable \texttt{LoRA} adapters. Let $\{h_i\}_{i=1}^n$ be the layers of the model to put the adapters on with pretrained matrices $\{W^i\}_{i=1}^n$. Then, the output of the layer $h_i$ when employing \texttt{WeightLoRA} is:
\begin{align}
  \label{eq:weightlora}  
  h_ix = W^ix + \omega_i \cdot A^i B^i x,
\end{align}
where $\omega_i$ represents the importance weight of the $i$-th adapter. The weights are composed into a vector $\omega = (\omega_1, \ldots, \omega_n)^T,$ which is then optimized throughout the fine-tuning process. To control the number of active adapters, we propose conditioning the $l_0$-``norm'' of $\omega$: $\|\omega\|_0 \overset{\text{def}}{=} \sum\nolimits_{i=1}^{n}\mathbb{I}(\omega_i \neq 0)$.

That is, if one uses the classical \texttt{LoRA} framework, the optimization problem to be solved is:
\begin{align*}
    \min_{A^i, B^i} \mathcal{L}(h_1, \ldots, h_n),
\end{align*}
where $\mathcal{L}$ denotes the loss function. We omit unnecessary dependencies in $\mathcal{L}$. In contrast, in our method, the optimization problem transforms into:
\begin{align}
    \label{eq:wloraproblem}
   \underset{s.t.\,\,\|\omega\|_0\, \leq\, K}{\min_\omega}\min_{A^i, B^i} \mathcal{L}(h_1, \ldots, h_n),
\end{align}
where $K$ is a positive constant defining the number of adapters to keep. The forward of the layer $h_i$ can now be rewritten as: 
\begin{equation*}
    h_ix = W^ix +
    \begin{cases}
        \omega_iA^i B^ix, \,\, \text{if}\,\, \omega_i > 0 \\
        0, \,\, \text{else} 
    \end{cases}
\end{equation*}
This is where the advantage of our method over the classical approach becomes clear. In fact, instead of optimizing a large number of matrices $A^i, B^i$, we switch to optimizing the vector $\omega$, which has a much smaller number of parameters. After $T$ training steps, we fix the vector $\omega$ and disconnect the adapters with $\omega_i = 0$, leading to training only the $K \le n$ most informative adapters.

\textbf{Discussion.} The sparsity constraint in the external minimization problem in \eqref{eq:weightlora} has been extensively studied \cite{buhlmann2011statistics, yuan2021stability, de2022zeroth, wu2024iterative}. Among others, we choose the \texttt{StoIHT} algorithm from \cite{nguyen2014linearconvergencestochasticiterative} due to its computational efficiency and simplicity in implementation. The idea behind this method is that after each $\omega$ gradient step, we retain only the Top-$K$ coordinates and nullify the others. In the literature, there are more sophisticated optimization methods for the zero norm constraint, such as \texttt{StoGradMP} \cite{nguyen2014linearconvergencestochasticiterative, li2016nonconvex, zhou2018efficient, shen2018tight}. However, all these methods require additional memory and computational time compared to the classical \texttt{StoIHT} algorithm. Also, experiments have shown that vanilla \texttt{StoIHT} is sufficient for achieving the performance of \texttt{WeightLoRA}.


It is important to note that simply adding the weights $\omega_i$ to the matrices $A^iB^i$ does not make much sense, given that their parameters are already trainable. That is why incorporating the $l_0$-norm is a fundamental component of our method. Due to the same trainability, the characteristic values of the weights $w_i$ are expected to be either $0$ or $1$, aligning with the observations made in numerical experiments (see Section \ref{sec:exp}). 

It is also important to acknowledge that while the number of adapters to tune $K$ and the number of steps after which unnecessary adapters are disconnected $T$ are considered hyperparameters, they can be selected based on considerations of available memory capacity and the required final accuracy of the model's performance. Specifically, the parameter $T$ does not exceed the duration of a single epoch and can be fixed to a relatively small number of steps. As for hyperparameter $K$, it is evident from a theoretical standpoint that it is optimal to maintain as many adapters as possible within the limit of sufficient training time. This increases the number of parameters, thus preserving the generalization capability. However, in practice, the value of $K$ is determined by the capacity of the fine-tuning device, the size of the training dataset, batch size, and the values of other hyperparameters. In experimental Section \ref{sec:exp}, we consistently observe that even 10\% of total \texttt{LoRA} heads are sufficient to attain optimal performance and less memory-consuming learning. 

We also emphasize that our approach reduces the number of trainable parameters and enhances the training process. When the timestamp $T$ is reached, the batch size can be increased as the number of parameters to store decreases significantly. This results in faster epochs and better convergence.

\subsection{\texttt{WeightLoRA+}}
In this section, we naturally extend the previous idea of selecting the most critical adapters by introducing \texttt{WeightLoRA+}.
\begin{table*}[!thbp]
\centering
\captionof{table}{Results of fine-tuning the DeBERTaV3-base model on GLUE benchmark. The best results are shown in \textbf{bold}, and the second-best are \underline{underlined}. 
The results are reported as the average of five runs with different random seeds. All experiment details are provided in Appendix \ref{appendix:deberta_glue}.
}
\resizebox{\textwidth}{!}{
\begin{tabular}{l|c|ccccccc|c}
\toprule
\multirow{2}*{\bf Method} & \multirow{2}*{\bf \small \# Params} & {\bf MNLI} & {\bf SST-2} & {\bf CoLA} & {\bf QQP } & {\bf QNLI} & {\bf RTE}  & {\bf MRPC}  & {\bf ALL} \\
~ & ~ & {Acc} & {Acc} & {Mcc} & {Acc/F1} & {Acc} & {Acc} & {Acc/F1} & {Avg} \\
\midrule 
{\texttt{Full Fine-Tuning}} & {184M (100\%)} & {0.8910} & {0.9541} & {0.6806} & {0.8962/0.8644} & {0.9383} & {0.8376} & {0.8848/0.9165} & {\bf 0.8689} \\ 
$\texttt{LoRA}_{r=8}$ \cite{hu2021lora} & {442K (0.24\%)} & {0.8797} & {0.9450} & {0.6913} & {0.8802/0.8437} & {0.9301} & {0.8448} & {0.8897/0.9220} & \textbf{0.8655} \\
\midrule
{$\texttt{AdaLoRA}$} \cite{zhang2023adalora}& {664K (0.36\%)} & {0.8390} & {0.9392} & {0.6222} & {0.8534/0.8083} & {0.8999} & {0.7834} & {0.7059/0.8225} & {0.8099} \\
$\texttt{IncreLoRA}$ \cite{zhang2023increlora} & {589K (0.32\%)} & {0.8046} & {0.8303} & {0.6266} & {0.8206/0.7938} & {0.8556} & {0.7426} & {0.7598/0.8435} & {0.7812} \\
$\texttt{Dynamic LoRA}$ \cite{mao2024dora} & {626K (0.34\%)} & {0.8545} & {0.9258} & {0.6490} & {0.8961/0.8643} & {0.9215} & {0.7315} & {0.8172/0.8701} & {0.8295} \\
{$\texttt{LoRA-drop}$} \cite{zhou2024lora}& {221K (0.12\%)} & {0.7659} & {0.9358} & {0.5903} & 0.8186/0.7701 & {0.8480} & {0.7668} & 0.7605/0.8283 & {0.7851} \\
$\texttt{ALoRA}$ \cite{liu2024alora} & {664K (0.36\%)} & {0.8301} & {0.9559} & {0.6419} & 0.{8408/0.8098} & {0.8833} & {0.7644} & 0.8572/0.9086 & {0.8263} \\
\texttt{MELoRA} \cite{ren2024melora} & \underline{110K (0.06\%)} & 
{0.7986} & {0.9358} & {0.6188} & 0.7599/0.8342 & {0.9170} & {0.7527} & 0.8138/0.8701 & {0.8088} \\
\texttt{FlexLoRA} \cite{wei2024flexora} & {442K (0.24\%)} & {0.7776} & {0.8204} & {0.6165} & {0.7959/0.7510} & {0.8383} & {0.7246} & {0.7187/0.8019} & {0.7587} \\
\midrule 
{$\texttt{WLoRA}_{r=8, k=1}$} & \textbf{12.3K (0.007\%)} & {0.7912} & {0.8876} & {0.5624} & {0.8307/0.7813} & {0.8724} & {0.7271} & {0.8309/0.8825} & {0.7862} \\
{$\texttt{WLoRA}_{r=8, k=5}$} & \textbf{61.5K (0.03\%)} & {0.8638} & {0.9427} & {0.6491} & {0.8685/0.8292} & {0.9136} & {0.7560} & {0.8799/0.9148} & \underline{0.8388} \\
{$\texttt{WLoRA}_{r=8, k=10}$} & \underline{123K (0.07\%)} & {0.8706} & {0.9473} & {0.6703} & {0.8728/0.8362} & {0.9167} & {0.7643} & {0.8775/0.9101} & \underline{0.8454} \\
\bottomrule
\end{tabular}
}
\label{table:glue_all_adapters}
\end{table*}
Formally, again let $\{h_i\}_{i = 1}^n$ be the layers of the model to put the adapters $\Delta W^i = A^iB^i$ on with pretrained matrices $\{W^i\}$, let $\omega = [\omega_1, \ldots, \omega_n]^T$ be the corresponding weights and $\{r_i\}_{i=1}^n$ be the corresponding ranks of $A^i$ and $B^i$. 

\textbf{Phase 1: Warm-up.} At the beginning of the fine-tuning process, we perform $T$ steps of \texttt{WeightLoRA} using an update rule of the form \eqref{eq:weightlora}. At this stage, the ranks $\{r_i\}_{i=1}^n$ remain constant.

\textbf{Phase 2: Expanded adapters training.} Next, we disconnect unnecessary LoRAs and increase the ranks of the rest in a way that the total memory consumption does not change: 
\begin{equation}
    h_i = W^ix +
    \begin{cases}
        \omega_i\tilde A^i\tilde B^i x, \,\, \text{if}\,\, \omega_i > 0 \\
        0, \,\, \text{else} 
    \end{cases},
\end{equation}
where $\tilde A^i = 
\begin{bmatrix}
    A^i\,\, A_{\text{ext}}^i
\end{bmatrix}, 
\tilde B^i = 
\begin{bmatrix}
    B^i\,\, B_{\text{ext}}^i 
\end{bmatrix}^T$, $A_{\text{ext}}^i \in \mathbb{R}^{d\times \tilde r_i}$, $B_{\text{ext}}^i \in \mathbb{R}^{\tilde r_i \times k}$.
These expanded matrices $\tilde A^i,\tilde B^i$ are then fine-tuned for better generalization. 

\textbf{Discussion.} 
In this setting, the central bottleneck is the choice of initialization of matrices $A_{\text{ext}}^i$ and $B_{\text{ext}}^i$. Consider \texttt{LoRA} adapter $A_i\cdot B_i\in \mathbb{R}^{d\times k}$ with $A_i \in \mathbb{R}^{d\times r_i}, B_i \in \mathbb{R}^{r_i\times k}$. We wish to extend this adapter to the rank $r^{\text{new}}_i > r_i$ and simultaneously ensure that the minimizer of the problem \eqref{eq:wloraproblem} does not shift to the point where all previous training becomes useless.  
To address this problem, we investigate several approaches:
 \begin{itemize}
    \item Extend existing matrices as it was done in the original paper \cite{hu2021lora}, i.e., $A$ is expanded via a random Gaussian initialization and $B$ is expanded with zeros:
    \begin{align*}
        \tilde A^i &= 
        \begin{bmatrix}
            A^i\,\, \mathcal{N}_{k\times (r^{\text{new}}_i - r_i)}
        \end{bmatrix},
        \\
        \\
        \tilde B^i &= 
        \begin{bmatrix}
            B^i \\ \mathbf{0}_{k\times (r^{\text{new}}_i - r_i)}
        \end{bmatrix}.
    \end{align*}
    \item Compute QR factorization of matrix $A_i$ with $Q_A \in \mathbb{C}^{d \times r_{i}}, R_A \in \mathbb{C}^{r_{i} \times r_{i}}$, it takes $O( d r_{i}^2)$ time.
        $
        A_{i} \cdot B_{i}^* = 
        Q_A R_A \cdot B^*
        $.
        Let us set extended matrices as
        \[
        \tilde A^i = \begin{bmatrix}
            Q_A & (I - Q_AQ_A^*) \mathcal{N}_{k\times (r^{\text{new}}_i - r_i)}
        \end{bmatrix}
        ,
        \]
        \[
        \tilde B^i = \begin{bmatrix}
            R_A B_i \\
            \mathbf{0}_{k\times (r^{\text{new}}_i - r_i)}
        \end{bmatrix}
        .
        \]
\end{itemize}
 Extending $A_i$ and $B_i$ in both cases does not change the adapter's value, thus preserving the impact of previous training.

\section{Experiments}
\label{sec:exp}
In this section, we provide a comprehensive experimental validation of our approach. We evaluate the fine-tuning performance of \texttt{WeightLoRA} and \texttt{WeightLoRA+} on DeBERTaV3-base \cite{he2021debertav3}, BART-large~\cite{lewis2019bart} and Llama3-7B \cite{grattafiori2024llama3herdmodels} models. Our experiments cover a wide range of tasks including natural language understanding with the GLUE benchmark~\cite{wang2018glue}, question answering with SQuADv1 \cite{rajpurkar2016squad} and SQuADv2 \cite{rajpurkar2018know} and natural language generation with XSum \cite{narayan2018don} and CNN/DailyMail \cite{hermann2015teaching} datasets. We provide the full experimental setup, additional details on hyperparameters, and the fine-tuning process in Appendix \ref{appendix:experiments}.


\subsection{Natural Language Understanding}
\label{sec:nlu}



In this section, we compare \texttt{WeightLoRA} with the methods employing dynamic rank allocation described in Section \ref{sec:related} on the GLUE benchmark \cite{wang2018glue}. 


\begin{table*}[htbp]
\centering
\caption{Results of fine-tuning the DeBERTaV3-base model on GLUE benchmark with more careful hyperparameter tuning, different ranks, and applying adapters to all attention layers. For every rank $r$, the best results on each task are shown in \textbf{bold}. All experiment details are provided in Appendix \ref{appendix:deberta_glue}.}
\resizebox{\textwidth}{!}{
\begin{tabular}{c|l|c|ccccccc|c}
\toprule
\multirow{2}*{\bf Rank} & \multirow{2}*{\bf Method} & \multirow{2}*{\bf \# Params} & {\bf MNLI} & {\bf SST-2} & {\bf CoLA} & {\bf QQP } & {\bf QNLI} & {\bf RTE}  & {\bf MRPC}  & {\bf ALL} \\
    &~ & ~ & {Acc} & {Acc} & {Mcc} & {Acc/F1} & {Acc} & {Acc} & {Acc/F1} & {Avg} \\
    \midrule 
\multirow{3}*{$r = 1$} & \texttt{LoRA} & 0.07\% & {0.8912} & {0.9488} & {0.6558} & {0.8905/0.8544} & {0.9250} & {0.7943} & {0.8905/0.9213} & {0.8562} \\
~ & $\texttt{WLoRA}_{k=20}$ & {\bf 0.02\%} & {0.8740} & {0.9431} & {0.6465} & {0.8813/0.8490} & {0.9254} & {0.7764} & {0.8882/0.9210} & {0.8479} \\ 
~ & $\texttt{WLoRA+}$ & 0.07\% & {0.8977} & {0.9495} & {0.6620} & {0.8841/0.8501} & {0.9246} & {0.8255} & {0.8873/0.9190} & {\bf 0.8612} \\ 
\midrule
\multirow{3}*{$r = 2$} & \texttt{LoRA} & 0.13\% & {0.9084} & {0.9413} & {0.6600} & {0.9077/0.8631} & {0.9369} & {0.8068} & {0.9003/0.9193} & {0.8641} \\ 
~ & $\texttt{WLoRA}_{k=20}$ & {\bf 0.04\%} & {0.9089} & {0.9518} & {0.6728} & {0.8848/0.8733} & {0.9048} & {0.7823} & {0.9004/0.9276} & {0.8591} \\ 
~ & $\texttt{WLoRA+}$ & 0.13\% & {0.8960} & {0.9524} & {0.6848} & {0.8898/0.8425} & {0.9149} & {0.8567} & {0.8910/0.9009} & {\bf 0.8667} \\ 
\midrule
\multirow{3}*{$r = 4$} & \texttt{LoRA} & 0.26\% & 
{0.8780} & {0.9562} & {0.6809} & {0.9153/0.8866} & {0.9582} & {0.7816} & {0.9207/0.9471} & {0.8700} \\ 
~ & $\texttt{WLoRA}_{k=20}$ & {\bf 0.09\%} & {0.8600} & {0.9614} & {0.6824} & {0.9005/0.8762} & {0.9200} & {0.8637} & {0.8901/0.9099} & {0.8680} \\ 
~ & $\texttt{WLoRA+}$ & 0.26\% & {0.9241} & {0.9714} & {0.6907} & {0.9032/0.9133} & {0.9057} & {0.8434} & {0.8615/0.8912} & {\bf 0.8743} \\
\midrule
\multirow{3}*{$r = 8$} & \texttt{LoRA} & 0.51\% & {0.9030} & {0.9641} & {0.6965} & {0.9092/0.8723} & {0.9449} & {0.8537} & {0.9064/0.9127} & {0.8804} \\
~ & $\texttt{WLoRA}_{k=20}$ & {\bf 0.17\%} & {0.9109} & {0.9653} & {0.6898} & {0.9130/0.8744} & {0.9467} & {0.7955} & {0.9061/0.9315} & {0.8744} \\ 
~ & $\texttt{WLoRA+}$ & 0.51\% & {0.9074} & {0.9644} & {0.6934} & {0.9212/0.8926} & {0.9466} & {0.8531} & {0.9023/0.9356} & {\bf 0.8844} \\
\bottomrule
\end{tabular}
}
    \label{tab:deberta_glue_new}
\end{table*}
\textbf{Dynamic rank comparison.} Initially, we perform a comparative analysis against \texttt{LoRA}-inspired methodologies that incorporate dynamic rank allocation: \texttt{AdaLoRA} \cite{zhang2023adalora}, \texttt{IncreLoRA} \cite{zhang2023increlora}, \texttt{Dynamic LoRA} \cite{mao2024dora}, \texttt{LoRA-drop} \cite{zhou2024lora}, \texttt{ALoRA} \cite{liu2024alora}, \texttt{MELoRA} \cite{ren2024melora}, and \texttt{FlexLoRA} \cite{wei2024flexora}. We employ the DeBERTaV3-base \cite{he2021debertav3} model, comprising 184M parameters. In this experimental setup, following the orginal paper, \texttt{LoRA} adapters are applied exclusively to the self-attention weights. This results in a configuration of $n=36$ fine-tuned layers. The motivation for this experiment is to evaluate existing adaptive layer selection methods in practice with minimal learning rate tuning (see Appendix \ref{appendix:deberta_glue} for more experimental details and hyperparameter tuning). The results of the experiment are shown in Table \ref{tab:deberta_glue_new}. 

The results indicate that all existing adaptive layer selection methods for fine-tuning require extensive hyperparameter tuning and perform poorly "out of the box", while the proposed \texttt{WeightLoRA} algorithm demonstrates competitive performance to standard \texttt{LoRA} and \texttt{Full FT} approaches.

\textbf{Tuning LoRA comparison.} Since the results of the previous experiment (Table \ref{tab:deberta_glue_new}) show that the dynamic rank allocation methods perform poorly and due to computational constraints, we do not consider these methods further and instead focus on the comparison with the classical \texttt{LoRA} algorithm. In the new experiment, we maintain the same setting as in the previous experiment but apply adapters to all attention layers (not just self-attention), resulting in $n=72$. We consider four values of \texttt{LoRA} adapter ranks for greater generalization, and we also perform extensive hyperparameter tuning, including a wide grid search for the learning rate and tuning of the warmup step in the linear scheduler (see Appendix \ref{appendix:deberta_glue} for details). The results of the experiment are presented in Table \ref{tab:deberta_glue_new}.

These results show that our proposed rank extension method, \texttt{WeightLoRA+}, performs better on average than \texttt{LoRA} for all ranks. At the same time, the adapter deactivation method, \texttt{WeightLoRA}, achieves results comparable in quality to the \texttt{LoRA} baseline while using approximately three times fewer parameters.

\textbf{LLama3 8B experiment.} In order to provide scalability of our results, we now consider a much larger Llama3-7B model for the same GLUE tasks. For this model, $n = 224$, and we compare the \texttt{LoRA} method with \texttt{WeightLoRA} using $K = 100$ and \texttt{WeightLoRA+}. The results of the experiment are presented in Table \ref{table:llama}. They show that for larger models, deactivating certain adapters can even lead to improved performance for higher ranks (see $r = 4, 8$ in Table \ref{table:llama}). Hyperparameter tuning is performed similarly to the previous experiment, with the main difference being the number of training steps (see Appendix \ref{appendix:deberta_glue} for details).

\begin{table*}[ht]
\centering
\caption{Results of fine-tuning the Llama3 8B model on GLUE benchmark. For every rank $r$, the best results on each task are shown in \textbf{bold}. All experiment details are provided in Appendix \ref{appendix:deberta_glue}.}
\label{table:llama}
\resizebox{\textwidth}{!}{
\begin{tabular}{c|l|c|cccccccc|c}
\toprule
\multirow{2}*{\bf Rank} & \multirow{2}*{\bf Method} & \multirow{2}*{\bf \# Params} & {\bf MNLI} & {\bf SST-2} & {\bf CoLA} & {\bf QQP} & {\bf QNLI} & {\bf RTE} & {\bf MRPC} & {\bf STS-B} & {\bf ALL} \\
 & & & {Acc} & {Acc} & {Mcc} & {Acc/F1} & {Acc} & {Acc} & {Acc/F1} & {Corr} & {Avg} \\
\midrule
\multirow{3}*{$r=1$} 
  & \texttt{LoRA} & 0.07\% & 0.9213 & 0.9703 & 0.6867 & 0.9001/0.8438 & 0.9208 & 0.8808 & 0.9406/0.9578 & 0.9212 & {\bf 0.8903} \\
  & $\texttt{WLoRA}_{k=100}$ & {\bf 0.03\%} & 0.8913 &  0.9604 & 0.6275 & 0.9001/0.8438 & 0.9208 & 0.8810 & 0.9406/0.9578 & 0.9653 & 0.8834 \\
  & $\texttt{WLoRA+}$ & 0.07\% & {0.9012} & {0.9682} & {0.6365} & {0.9100/0.8522} & {0.9304} & {0.8897} & {0.9491/0.9657} & {0.9737} & {\bf 0.8923} \\
\midrule
\multirow{3}*{$r=2$} 
  & \texttt{LoRA} & 0.13\% & 0.9222 & 0.9703 & 0.6867 & 0.9001/0.8485 & 0.9209 & 0.8911 & 0.9409/0.9598 & 0.9220 &  0.8922 \\
  & $\texttt{WLoRA}_{k=100}$ & {\bf 0.07\%} & 0.9012 & 0.9703 & 0.6475 & 0.9001/0.8485 & 0.9209 & 0.9109 & 0.9408/0.9593 & 0.9687 & 0.8930 \\
  & $\texttt{WLoRA+}$ & 0.13\% & {0.9133} & {0.9703} & {0.6587} & {0.9118/0.8601} & {0.9330} & {0.9234} & {0.9510/0.9695} & {0.9806} & {\bf 0.9046} \\
\midrule
\multirow{3}*{$r=4$} 
  & \texttt{LoRA} & 0.26\% & 0.9301 & 0.9802 & 0.6913 & 0.9101/0.8585 & 0.9408 & 0.9109 & 0.9506/0.9678 & 0.9223 &  0.9024 \\
  & $\texttt{WLoRA}_{k=100}$ & {\bf 0.13\%} & 0.9111 & 0.9802 & 0.6784 & 0.9208/0.8752 & 0.9506 & 0.9110 & 0.9507/0.9696 & 0.9687 & 0.9073 \\
  & $\texttt{WLoRA+}$ & 0.26\% & {0.9213} & {0.9802} & {0.6887} & {0.9312/0.8873} & {0.9607} & {0.9223} & {0.9621/0.9804} & {0.9810} & {\bf 0.9168} \\
\midrule
\multirow{3}*{$r=8$} 
  & \texttt{LoRA} & 0.52\% & 0.9447 & 0.9802 & 0.7015 & 0.9111/0.8658 & 0.9507 & 0.9110 & 0.9557/0.9617 & 0.9290 & 0.9080 \\
  & $\texttt{WLoRA}_{k=100}$ & {\bf 0.26\%} & 0.9201 & 0.9802 & 0.6885 & 0.9208/0.8757 & 0.9604 & 0.9208 & 0.9604/0.9718 & 0.9696 & 0.9130 \\
  & $\texttt{WLoRA+}$ & 0.52\% & {0.9342} & {0.9802} & {0.7057} & {0.9324/0.8912} & {0.9771} & {0.9312} & {0.9740/0.9879} & {0.9854} & {\bf 0.9258} \\
\bottomrule
\end{tabular}
}
\end{table*}

The experimental results in this section demonstrate the broad applicability of the \texttt{WeightLoRA} method for models with different numbers of trainable parameters. They also highlight the advantages of the alternative adapter weight optimization approach used in \texttt{WeightLoRA} over existing dynamic weight allocation methods.

\subsection{Question Answering}
\label{sec:qa}
In this experiment, we consider the more challenging question-answering task on SQuAD v1.1 \cite{rajpurkar2016squad} and SQuAD v2.0 \cite{rajpurkar2018know} datasets. This requires solving a task—specific sequence labeling, predicting the probability that each token represents the start or end of the answer span. We refer to the Appendix \ref{appendix:deberta_squad} for a detailed description on experimental setup. The results are summarized in Table \ref{tab:squad_experiments}. 

\begin{table}[ht]
\centering
\caption{Results of fine-tuning DeBERTaV3-base on SQuAD v1.1 and SQuAD v2.0 datasets. We report F1-scores averaged by $5$ runs. The best results are shown in \textbf{bold}, the second-best are \underline{underlined}. All experiment details are provided in Appendix \ref{appendix:deberta_squad}.}
\resizebox{\textwidth}{!}{
\begin{tabular}{ll|cccc|cccc}
\toprule
 & & \multicolumn{4}{c|}{\bf SQuADv1.1} & \multicolumn{4}{c}{\bf SQuADv2.0} \\
\cline{3-10} 
{\bf Method} &{\bf Metric} & {${r=1}$} & {${r=2}$} & {${r=4}$} & {${r=8}$} & {${r=1}$} & {${r=2}$} & {${r=4}$} & {${r=8}$} \\
\midrule
\texttt{LoRA}& F1-score & {\underline{90.01}} & {\underline{92.05}} & {\underline{92.17}} & {\underline{92.29}} & {80.19} & {\underline{82.98}} & {\underline{84.40}} & {\underline{85.19}} \\
& \# Params & {0.07\%} & {0.13\%} & {0.26\%} & {0.51\%} & {0.07\%} & {0.13\%} & {0.26\%} & {0.51\%} \\
\midrule
{$\texttt{WLoRA}_{k=20}$}& F1-score & {89.12} & {89.12} & {92.08} & {92.07} & {\underline{80.37}} & {82.73} & {84.23} & {\underline{85.19}} \\
&\# Params & {\bf 0.02\%} & {\bf 0.04\%} & {\bf 0.09\%} & {\bf 0.17\%} & {\bf 0.02\%} & {\bf 0.04\%} & {\bf 0.09\%} & {\bf 0.17\%} \\
\midrule
{$\texttt{WLoRA}+_{k=15}$}& F1-score & {\bf 91.37} & {\bf 92.20} & {\bf 92.48} & {\bf 92.61} & {\bf 80.65} & {\bf 83.13} & {\bf 84.86} & {\bf 85.49} \\
& \# Params & {0.07\%} & {0.13\%} & {0.26\%} & {0.51\%} & {0.07\%} & {0.13\%} & {0.26\%} & {0.51\%} \\
\bottomrule
\end{tabular}
}
\label{tab:squad_experiments}
\end{table}

Analysis of these results indicates that the proposed \texttt{WeightLoRA+} method outperforms the classical \texttt{LoRA} approach for every rank selection. At the same time, the \texttt{WeightLoRA} method demonstrates competitive performance compared to both methods. It achieves this with a number of trainable parameters reduced by a factor of three. 

\subsection{Natural Language Generation}
\label{sec:nlg}
In this section, we address the task of natural language generation (NLG). We apply the \texttt{WeightLoRA} method to fine-tune a BART-large model \cite{lewis2019bart} on two widely used NLG benchmarks: XSum \cite{narayan2018don} and CNN/DailyMail \cite{hermann2015teaching}. The XSum benchmark focuses on abstract single-document summarization, where the objective is to generate a concise one-sentence summary. 
The CNN/DailyMail dataset consists of news stories from the CNN and Daily Mail websites, paired with human-generated summary bullets. The summaries are presented as fill-in-the-blank questions (with one entity hidden), and the corresponding news stories serve as passages from which the system is expected to generate the answer. 

\begin{table}[ht]
\centering
\caption{Results of fine-tuning BART-large on XSum and CNN/DailyMail datasets. We report rogue1-scores averaged by $5$ runs. The best results are shown in \textbf{bold}, the second-best are \underline{underlined}. All experiment details are provided in Appendix \ref{appendix:deberta_squad}.}
\resizebox{\textwidth}{!}{
\begin{tabular}{ll|cccc|cccc}
\toprule
& & \multicolumn{4}{c|}{\bf XSum} & \multicolumn{4}{c}{\bf CNN/DailyMail} \\
\cline{3-10} 
{\bf Method} &{\bf Metric} & {$r=1$} & {$r=2$} & {$r=4$} & {$r=8$} & {$r=1$} & {$r=2$} & {$r=4$} & {$r=8$} \\
\midrule
{\texttt{LoRA}} & F1-score & {35.23} & {35.59} & {35.96} & {36.07} & {\underline{24.43}} & {\underline{24.50}} & {\underline{25.16}} & {\underline{25.28}} \\
& \# Params & {0.13\%} & {0.27\%} & {0.53\%} & {1.05\%} & {0.13\%} & {0.27\%} & {0.53\%} & {1.05\%} \\
\midrule
{$\texttt{WLoRA}_{k=20}$}& F1-score & {\underline{35.63}} & {\underline{36.11}} & {\underline{36.27}} & {\underline{36.56}} & {\underline{24.43}} & {24.43} & {24.67} & {24.96} \\
&\# Params & {\bf 0.01\%} & {\bf 0.03\%} & {\bf 0.06\%} & {\bf 0.11\%} & {\bf 0.01\%} & {\bf 0.03\%} & {\bf 0.06\%} & {\bf 0.11\%} \\
\midrule
{$\texttt{WLoRA+}_{k=48}$} & F1-score & {\bf 35.82} & {\bf 36.37} & {\bf 36.45} & {\bf 36.74} & {\bf 24.45} & {\bf 24.72} & {\bf 25.31} & {\bf 25.72} \\
& \# Params & {0.13\%} & {0.27\%} & {0.53\%} & {1.05\%} & {0.13\%} & {0.27\%} & {0.53\%} & {1.05\%} \\
\bottomrule
\end{tabular}
}
\label{table:nlg}
\end{table}
\begin{table*}[!t]
\centering
\caption{Ablation study about random adapters disabling.
The results are reported as the average of five runs with different random seeds. The best results are shown in \textbf{bold}.
}
\resizebox{\textwidth}{!}{
\begin{tabular}{c|l|ccccccc|c}
\toprule
\multirow{2}*{\bf \# Active Adapters} & \multirow{2}*{\bf Method} & {\bf MNLI} & {\bf SST-2} & {\bf CoLA} & {\bf QQP } & {\bf QNLI} & {\bf RTE}  & {\bf MRPC}  & {\bf ALL} \\
~ & ~ & {Acc} & {Acc} & {Mcc} & {Acc/F1} & {Acc} & {Acc} & {Acc/F1} & {Avg} \\
\midrule 
\multirow{2}*{$K = 1$} & {$\texttt{WLoRA}$} & {0.7912} & {0.8876} & {0.5624} & {0.8307/0.7813} & {0.8724} & {0.7271} & {0.8309/0.8825} & {\bf 0.7862} \\
~ & {$\texttt{RLoRA}$} & {0.5124} & {0.4908} & {0.0115} & {0.6318/0.4265} & {0.4946} & {0.5271} & {0.3162/0.3532} & {0.4143} \\
\midrule 
\multirow{2}*{$K = 5$}  & {$\texttt{WLoRA}$} & {0.8638} & {0.9427} & {0.6491} & {0.8685/0.8292} & {0.9136} & {0.7560} & {0.8799/0.9148} & \textbf{0.8388}  \\
~ & {$\texttt{RLoRA}$} & {0.7260} & {0.9178} & {0.5217} & {0.8396/0.8001} & {0.8640} & {0.7160} & {0.8440/0.8905} & {0.7761} \\
\midrule 
\multirow{2}*{$K = 10$}  & {$\texttt{WLoRA}$} &  {0.8706} & {0.9473} & {0.6703} & {0.8728/0.8362} & {0.9167} & {0.7643} & {0.8775/0.9101} & \textbf{0.8454} \\
~ & {$\texttt{RLoRA}$} & {0.8209} & {0.9300} & {0.5368} & {0.8539/0.8110} & {0.8875} & {0.7316} & {0.8529/0.8950} & {0.8019} \\
\bottomrule
\end{tabular}
}
\label{table:ablation}
\end{table*}

The experimental results are presented in Table \ref{table:nlg}. The results demonstrate that \texttt{WeightLoRA+} consistently achieves superior performance compared to the standard \texttt{LoRA} method when compared with the same number of trainable parameters. Second-best results are typically found to be obtained by \texttt{WeightLoRA}, despite the fact that the number of its trainable parameters is significantly smaller.

%
Accordingly, these results suggest that our approach is consistently superior to the classical one in that comparable results are obtained for an algorithm with significantly fewer parameters and better results are obtained for an algorithm with a comparable number of parameters.

\subsection{Ablation Study}
\label{appendix:ablation}

In this section, we show that the selection of adapters by rule \eqref{eq:wloraproblem} is essential in the \texttt{WeightLoRA} technique. We implement exactly the same method as \texttt{WeightLoRA}, except that we do not disable the adapters by an optimized weights $\omega$, but randomly. In this paper, we named this method \texttt{RLoRA}. The results comparing \texttt{WeightLoRA} and \texttt{RLoRA} in an experiment similar to Table \ref{table:glue_all_adapters} from Section \ref{sec:nlu} with fixed rank $r=8$ are summarized in Table \ref{table:ablation}. 

The result of this experiment shows that the random disabling of the adapters performs poorly. This again confirms that optimizing the weights $\omega$ is necessary for the precise performance of our method \texttt{WeightLoRA}.
\section{Conclusion}
\label{sec:conclusion}
In this paper, we introduce the \texttt{WeightLoRA} framework, which adaptively selects the most critical \texttt{LoRA} adapters during the training process. We propose two novel methods \texttt{WeightLoRA} and \texttt{WeightLoRA+} to advance the landscape of PEFT techniques for LLMs, offering a practical and effective solution to the growing demands of AI applications. We demonstrate that the use of our algorithms can significantly enhance the performance of PEFT for various tasks including natural language understanding, question-answering and natural language generation, and different pretrained models. 
The source code to reproduce the results is available\footnote{\url{https://github.com/brain-mmo-lab/WLoRA}}.





\end{mainpart}
\begin{appendixpart}
\section{\texttt{LoRA} Notation}
\label{sec:lora_notation}
Let us briefly describe the original \texttt{LoRA} \cite{hu2021lora} framework. Given a pre-trained weight matrix $W \in \mathbb{R}^{d\times k}$, \texttt{LoRA} aims to learn an additive update $\Delta W \in \mathbb{R}^{d\times k}$, which is constrained to a low-rank decomposition of the form: $\Delta W = AB$, where $A \in \mathbb{R}^{d\times r}$ and $B \in \mathbb{R}^{r \times k}$. It is assumed that the rank $r$ is relatively small, $r \ll \min(d, k)$, to reduce the number of trainable parameters. This low-rank decomposition enables task-specific information to be encoded in the matrices $A$ and $B$ without requiring fine-tuning of the entire $W$. Formally, given an input $x$, the forward of the layer $h$ is as follows:
\begin{equation}
    \label{eq:LoRA} 
    hx = Wx + \Delta Wx = Wx + ABx.
\end{equation}
In this paper, we also define the sets of all adapters $A$ and $B$ (for all layers) as $\mathcal{A}$ and $\mathcal{B}$, respectively.


\section{Experimental Details}
\label{appendix:experiments}

We use PyTorch \cite{paszke2019pytorchimperativestylehighperformance} to implement our algorithms and the publicly available implementations of the baselines. Our code is based on the publicly available Huggingface Transformers3 \cite{wolf2020huggingfacestransformersstateoftheartnatural} library. All the experiments are conducted on NVIDIA V100 16GB and NVIDIA A100 80GB GPU. All the results are averaged across $5$ random seeds.

In our experiments, all \texttt{LoRA}-like adapters are scaled by $ \alpha/r $ to achieve better training \cite{hu2021lora}. The hyperparameter $\alpha$ typically is set as $32$ and never  tuned \cite{zhang2023adalora, yang2020feature}. In all our experiments we also use the \texttt{LoRA}-dropout technique \cite{lin2024lora} with probability $p = 0.05$ to avoid overfitting.

For every experiment, we choose default values of non-tuned hyperparameters of the Adam optimizer, such as $\beta_1, \beta_2 = 0.9, 0.999$. 

\subsection{Additional Details about NLU Experiment from Section \ref{sec:nlu}}
\label{appendix:deberta_glue}

\begin{table}[htbp]
    \centering
    \caption{Hyper-parameter setup for DeBERTa model on GLUE benchmark.}
    \begin{tabular}{l|cc}
    \toprule
    \textbf{Task} & \textbf{\# Epochs} & \textbf{Batch size}
    \\
    \midrule 
    {\bf MNLI} & $7$ & $ 32 $ 
    \\
    {\bf SST-2} & $10$ & $ 32 $
    \\
    {\bf CoLA} & $15$ & $ 64 $
    \\
    {\bf QQP } & $5$ & $32$ 
    \\
    {\bf QNLI} & $5$ & $ 32 $
    \\
    {\bf RTE} & $50$ & $ 64 $
    \\
    {\bf MRPC} & $30$ & $ 64 $
    \\
    \bottomrule
    \end{tabular}
    \label{table:app_glue_setup}
\end{table}
\textbf{Dynamic rank comparison.} We select the learning rate that achieves the best performance from the grid $\{5 \cdot 10^{-5},\ 5\cdot 10^{-4},\ 5 \cdot 10^{-3}\}$. We apply \texttt{LoRA}-like adapters only to the self-attention layers, while the other attention layers remain frozen. The total number of adapters for DeBERTaV3-base is $n = 36$.
For all GLUE tasks, we use one training epoch, a batch size of $32$, and $6$ gradient accumulation steps. The scheduler uses a linear schedule with $100$ warm-up steps. 

\textbf{Dynamic rank comparison.} We extend the learning rate grid from the previous experiment and perform tuning over the set $\{ 10^{-6},\ 3\cdot 10^{-5},\ 5 \cdot 10^{-5},\ 8 \cdot 10^{-5},\ 10^{-4},\ 3 \cdot 10^{-4},\ 5\cdot 10^{-4},\ 8\cdot 10^{-4},\ 10^{-3},\ 5 \cdot 10^{-3} \}$. For the second experiment, we tune the number of warmup steps from the grid $\{ 0,\ 20,\ 50,\ 70,\ 100,\ 200 \}$. The number of epochs and batch size for each task are provided in Table \ref{table:app_glue_setup}.

\textbf{LLama3 8B experiment.} We tune the learning rate using the same grid as in the previous experiment. Warmup steps were tuned on the set $\{10, 25, 30, 50\}$. The number of training steps and the batch size are adjusted according to the model size. In this experiment, we also use $2$ gradient accumulation steps instead of $6$, as was done in both GLUE experiments. The batch sizes and number of training steps for each task are provided in Table \ref{table:app_llama_setup}.

\begin{table}[h!]
\centering
\caption{Hyper-parameter setup for Llama 3 model GLUE benchmark.}
\begin{tabular}{l|cc}
\toprule
\textbf{Task} & \textbf{\# Training steps} & \textbf{Batch size}
\\
\midrule 
{\bf MNLI} & $256$ & $ 16 $ 
\\
{\bf SST-2} & $128$ & $ 32 $
\\
{\bf CoLA} & $1024$ & $ 16 $
\\
{\bf QQP } & $1024$ & $8$ 
\\
{\bf QNLI} & $256$ & $ 16 $
\\
{\bf RTE} & $256$ & $ 16 $
\\
{\bf MRPC} & $512$ & $ 32 $
\\
{\bf STS-B} & $256$ & $ 16 $
\\
\bottomrule
\end{tabular}
\label{table:app_llama_setup}
\end{table}

\subsection{Additional Details about QA and NLG Experiment from Sections \ref{sec:qa} and \ref{sec:nlg}}
\label{appendix:deberta_squad}


For both experiments from Sections \ref{sec:qa} and \ref{sec:nlg}, we perform Adam learning rate tuning from the grid $\{ 10^{-6}, 3\cdot 10^{-5}, 5 \cdot 10^{-5}, 8 \cdot 10^{-5}, 10^{-4}, 3 \cdot 10^{-4}, 5\cdot 10^{-4}, 8\cdot 10^{-4}, 10^{-3} \}$ and warm-up steps from the grid $\{ 0, 20, 50, 70, 100, 200 \}$. Other hyperparameters are shown in Table \ref{table:app_squad_setup}.

\begin{table}[!h]
\centering
\caption{Hyper-parameter setup for SQuADv1, v2, XSum and CNN/DailyMail benchmarks.}
\begin{tabular}{l|cccccc}
\toprule
\textbf{Dataset} & \textbf{\# epochs} & \textbf{batch size}
\\
\midrule 
{\bf SQuAD v1.1 } & $10$ & $ 32 $
\\
{\bf SQuADv2.0 } & $15$ & $32$
\\
{\bf XSum } & $35$ & $8$ &
\\
{\bf CNN/DailyMail } & $25$ & $8$ 
\\ 
\bottomrule
\end{tabular}
\label{table:app_squad_setup}
\end{table}

\end{appendixpart}
\end{document}